\def\year{2019}\relax
\documentclass[letterpaper]{article} %DO NOT CHANGE THIS
\usepackage{aaai19}  %Required
\usepackage{times}  %Required
\usepackage{helvet}  %Required
\usepackage{courier}  %Requiredhttps://preview.overleaf.com/public/fgkvvmtqyydg/images/8861f260b26cc96bce1a62d8c69ffc1108a25859.jpeg
\usepackage{url}  %Required
\usepackage{graphicx}  %Required

\usepackage{booktabs}       % professional-quality tables
\usepackage{amsfonts}       % blackboard math symbols
\usepackage{nicefrac}       % compact symbols for 1/2, etc.
\usepackage{microtype}      % microtypography
\usepackage{amsmath}
\usepackage{float}
\usepackage{color}
\usepackage{subfigure}
\usepackage{enumitem}
\DeclareMathOperator*{\argmin}{arg\,min}

\begin{document}
\title{Resisting Adversarial Attacks Using Gaussian Mixture Variational Autoencoders}
\author{  Partha Ghosh\thanks{Equal contribution} \\
  Max Planck Institute of Intelligent Systems\\
  T\"{u}bingen, Germany\\
%   \texttt{partha.ghosh@tuebingen.mpg.de} \\
  %% examples of more authors
  \And
  Arpan Losalka$^*$\\
  IBM Research AI\\
  New Delhi, India \\
%   \texttt{arlosalk@in.ibm.com} \\
  \And
  Michael J Black \\
  Max Planck Institute of Intelligent Systems\\
  T\"{u}bingen, Germany\\
%   \texttt{black@tue.mpg.de} \\
}
% \author{Anonymous submission to AAAI Press\\
% Association for the Advancement of Artificial Intelligence\\
% 2275 East Bayshore Road, Suite 160\\
% Palo Alto, California 94303\\
% }
\maketitle
\begin{abstract}
Susceptibility of deep neural networks to adversarial attacks poses a major theoretical and practical challenge. All efforts to harden classifiers against such attacks have seen limited success till now. Two distinct categories of samples against which deep neural networks are vulnerable, ``adversarial samples" and ``fooling samples", have been tackled separately so far due to the difficulty posed when considered together. In this work, we show how one can defend against them both under a unified framework. Our model has the form of a variational autoencoder with a Gaussian mixture prior on the latent variable, such that each mixture component corresponds to a single class. We show how selective classification can be performed using this model, thereby causing the adversarial objective to entail a conflict. The proposed method leads to the rejection of adversarial samples instead of misclassification, while maintaining high precision and recall on test data. It also inherently provides a way of learning a selective classifier in a semi-supervised scenario, which can similarly resist adversarial attacks. We further show how one can reclassify the detected adversarial samples by iterative optimization.\footnote{Accepted for publication in the Proceedings of the Thirty-Third AAAI Conference on Aritificial Intelligence (AAAI 2019)}
\end{abstract}

\section{Introduction}
\label{intro}
The vulnerability of deep neural networks to adversarial attacks has generated a lot of interest and concern in the past few years. The fact that these networks can be easily fooled by adding specially crafted noise to the input, such that the original and modified inputs are indistinguishable to humans \shortcite{szegedy2013intriguing}, clearly suggests that they fail to mimic the human learning process. Even though these networks achieve state-of-the-art performance, often surpassing human level performance \shortcite{he2015delving,huang2017densely} on the test data used for different tasks, their vulnerability is a cause of concern when deploying them in real life applications, especially in domains such as health care \shortcite{finlayson2018adversarial}, autonomous vehicles \shortcite{evtimov2017robust} and defense, etc.

\subsection{Adversarial Attacks and Defenses}
Adversarially crafted samples can be classified into two broad categories, namely (i) adversarial samples \shortcite{szegedy2013intriguing} and (ii) fooling samples as defined by \shortcite{nguyen2015deep}. Existence of adversarial samples was first shown by Szegedy et al.~\shortcite{szegedy2013intriguing}, while fooling samples \shortcite{nguyen2015deep}, which are closely related to the idea of ``rubbish class'' images \shortcite{lecun1998gradient} were introduced by Nguyen et al.~\shortcite{nguyen2015deep}. Evolutionary algorithms were applied to inputs drawn from a uniform distribution, using the predicted probability corresponding to the targeted class as the fitness function \shortcite{nguyen2015deep} to craft such fooling samples. It has also been shown that Gaussian noise can be directly used to trick classifiers into predicting one of the output classes with very high probability \shortcite{goodfellow2014explaining}.\\
\indent Adversarial attack methods can be classified into (i) white box attacks \shortcite{szegedy2013intriguing,goodfellow2014explaining,carlini2017towards,papernot2016limitations,moosavi2016deepfool,madry2017towards}, which use knowledge of the machine learning model (such as model architecture, loss function used during training, etc.) for crafting adversarial samples, and (ii) black box attacks \shortcite{papernot2017practical,papernot2016transferability,chen2017zoo}, which only require the model for obtaining labels corresponding to input samples. Both these kinds of attacks can be further split into two sub categories, (i) targeted attacks, which trick the model into producing a chosen output, and (ii) non-targeted attacks, which cause the model to produce any undesired output \shortcite{goodfellow2014explaining}. The majority of attacks and defenses have dealt with adversarial samples so far \shortcite{szegedy2013intriguing,gu2014towards,papernot2016distillation}, while a relatively smaller literature deals with fooling samples \shortcite{nguyen2015deep}. However, to the best of our knowledge, no prior method tries to defend against both kinds of samples simultaneously under a unified framework. State-of-the-art defense mechanisms have tried to harden a classifier by one or more of the following techniques: adversarial retraining \shortcite{szegedy2013intriguing}, preprocessing inputs \shortcite{gu2014towards}, deploying auxiliary detection networks \shortcite{Meng:2017:MTD:3133956.3134057} or obfuscating gradients \shortcite{obfuscated-gradients}. One common drawback of these defense mechanisms is that they do not eliminate the vulnerability of deep networks altogether, but only try to defend against previously proposed attack methods. Hence, they have been easily broken by stronger attacks, which are specifically designed to overcome their defense strategies \shortcite{carlini2016defensive,obfuscated-gradients}. 

Szegedy et al.~\shortcite{szegedy2013intriguing} argue that the primary reason for the existence of adversarial samples is the presence of small ``pockets'' in the data manifold, which are rarely sampled in the training or test set. On the other hand, Goodfellow et al.~\shortcite{goodfellow2014explaining} have proposed the ``linearity hypothesis'' to explain the presence of adversarial samples. Under our approach as detailed in Sec.~\ref{res_adv_smpls}, the adversarial objective poses a fundamental conflict of interest, and inherently addresses both these possible explanations.

\subsection{Approach} % Need a better name or needs to be deleted
\begin{figure*}[t]
  \begin{center}
  	\includegraphics[width=0.8\textwidth]{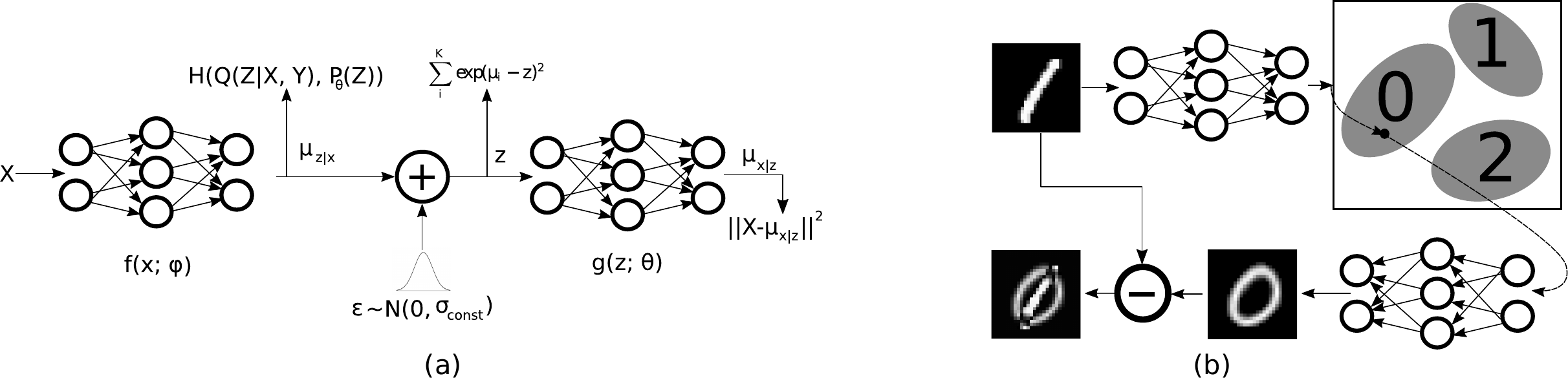}
  \end{center}
  \caption{(a) The model at training time. All the inputs are in green, while all the losses are in brown. (b) Model pipeline at inference time. The red dot shows that the attacker is successful in fooling the encoder by placing its output in the wrong class. However, it results in a high reconstruction error, since the decoder generates an image of the target class.}
  \label{model_pipeline}
\end{figure*}

We design a generative model that finds a latent random variable $\mathbf{z}$ such that data label $\mathbf{y}$ and the data $\mathbf{x}$ become conditionally independent given $\mathbf{z}$, i.e., $P(\mathbf{x}, \mathbf{y} | \mathbf{z}) = P(\mathbf{x}|\mathbf{z})P(\mathbf{y}|\mathbf{z})$. We base our generative model on VAEs \shortcite{kingma2013auto}, and obtain an inference model that represents $P(\mathbf{z}|\mathbf{x})$ and a generative model that represents $P(\mathbf{x}|\mathbf{z})$. We perform label inference $P(\mathbf{y}|\mathbf{x})$ by computing $\arg\max_{\mathbf{y}} P(\mathbf{y}|\mathbf{z}= \mathbb{E}_{P(\mathbf{z}|\mathbf{x})}[\mathbf{z}])$. We choose the latent space distribution $P(\mathbf{z})$ to be a  mixture of Gaussians, such that each mixture component represents one of the classes in the data. Under this construct, inferring the label given latent encoding, i.e., $P(\mathbf{y}|\mathbf{z})$ becomes trivial by computing the contribution of the mixture components. Adversarial samples are dealt with by thresholding in the latent and output spaces of the generative model and rejecting the inputs for which $P(\mathbf{x}) \approx 0$. In Figure \ref{model_pipeline}, we describe our network at test and train time.

Our contributions can be summarized as follows.
\begin{itemize}
% [leftmargin=0.75cm]
% \vspace{-0.15cm}
% \setlength\itemsep{-.1cm}
\item We show how VAE's can be trained with labeled data, using a Gaussian mixture prior on the latent variable in order to perform classification.
\item We perform selective classification using this framework, thereby rejecting adversarial and fooling samples.
\item We propose a method to learn a classifier in a semi-supervised scenario using the same framework, and show that this classifier is also resistant against adversarial attacks.
\item We also show how the detected adversarial samples can be reclassified into the correct class by iterative optimization.
\item We verify our claims through experimentation on 3 publicly available datasets: MNIST \shortcite{lecun1998gradient}, SVHN \shortcite{netzer2011reading} and COIL-100  \shortcite{nayar1996columbia}.
\end{itemize}

\section{Related Work}
A few pieces of work in the existing literature on defense against adversarial attacks have attempted to use generative models in different ways.

Samangouei et al. \shortcite{samangouei2018defensegan} propose training a Generative Adversarial Network (GAN) on the training data of a classifier, and use this network to project every test sample on to the data manifold by iterative optimization. This method does not try to detect adversarial samples, and does not tackle ``fooling images''. Further, this defense technique has been recently shown to be ineffective \shortcite{obfuscated-gradients}. Other pieces of work have also shown that adversarial samples can lie on the output manifold of generative models trained on the training data for a classifier \shortcite{zhao2017generating}.

PixelDefend, proposed by Song et al. \shortcite{song2017pixeldefend} also uses a generative model to detect adversarial samples, and then rectifies the classifier output by projecting the adversarial input back to the data manifold. However, Athalye et al. have shown that this method can also be broken by bypassing the exploding/vanishing gradient problem introduced by the defense mechanism.

MagNet \shortcite{meng2017magnet} uses autoencoders to detect adversarial inputs, and is similar to our detection mechanism in the way reconstruction threshold is used for detecting adversarial inputs. This defense method does not claim security in the white box setting. Further, the technique has also been broken in the grey box setting by recently proposed attack methods \shortcite{carlini2017magnet}. 

Traditional autoencoders do not constrain the latent representation to have a specific distribution like variational autoencoders. Our use of variational autoencoders allows us to defend against adversarial and fooling inputs simultaneously, by using thresholds in the latent and output spaces of the model in conjunction. This makes the method secure to white box attacks as well, which is not the case with MagNet.

Further, even state of the art defense mechanisms \shortcite{madry2017towards} and certified defenses have been shown to be ineffective for simple datasets such as MNIST \shortcite{song2018generative}. We show via extensive experimentation on different datasets how our method is able to defend against strong adversarial attacks, as well as end to end white box attacks.

\section{Method}

\subsection{Variational Autoencoders}

We consider the dataset $\mathbf{X}=\{\mathbf{x}^{(i)}\}_{i=1}^{N}$ consisting of $N$ i.i.d. samples of a random variable $\mathbf{x}$ in the space $\mathcal{X}$. Let $\mathbf{z}$ be the latent representation from which the data is assumed to have been generated. Similar to Kingma et al.~\shortcite{kingma2013auto}, we assume that the data generation process consists of two steps: (i) a value $\mathbf{z}^{(i)}$ is sampled from a prior distribution $P_{\theta^{*}}(\mathbf{z})$; (ii) a value $\mathbf{x}^{(i)}$ is generated from a conditional distribution $P_{\theta^{*}}(\mathbf{x|z})$. We also assume that the prior $P_{\theta^{*}}(\mathbf{z})$ and likelihood $P_{\theta^{*}}(\mathbf{x|z})$ come from parametric families of distributions $P_{\theta}(\mathbf{z})$ and $P_{\theta}(\mathbf{x|z})$ respectively. In order to maximize the data likelihood $P_{\theta}(\mathbf{x}) = \int P_{\theta}(\mathbf{z})P_{\theta}(\mathbf{x}|\mathbf{z})d\mathbf{z}$, VAEs \shortcite{kingma2013auto} use an encoder network $Q_\phi(\mathbf{z}|\mathbf{x})$, that approximates $P_\theta(\mathbf{z}|\mathbf{x})$. The evidence lower bound (ELBO) for VAE is given by
\begin{equation}
\begin{split}
ELBO(\mathbf{x},\theta,\phi) = \mathbb{E}_{\mathbf{z}\sim Q_{\phi}(\mathbf{z}|\mathbf{x})}[\log P_\theta(\mathbf{x}|\mathbf{z})]\\
- D_{\mathit{KL}}[Q_{\phi}(\mathbf{z}|\mathbf{x}) || P(\mathbf{z})]
\end{split}
\label{vae_loss}
\end{equation}
where $D_{\mathit{\mathit{KL}}}$ represents the KL divergence measure. Using a Gaussian prior $P_{\theta}(\mathbf{z})$ and a Gaussian posterior $Q_{\phi}(\mathbf{z}|\mathbf{x})$, variational autoencoders maximize this lower bound deriving a closed form expression for the KL divergence term.

\subsection{Modifying the Evidence Lower Bound}
VAEs do not enforce any lower or upper bound on encoder entropy $H(Q_\phi(\mathbf{z}|\mathbf{x}))$. This can result in blurry reconstruction due to sample averaging in case of overlap in the latent space. On the other hand, unbounded decrease in $H(Q_\phi(\mathbf{z}|\mathbf{x}))$ is not desirable either, as in that case the VAE can degenerate to a deterministic autoencoder leading to holes in the latent space. Hence, we seek an alternative design in which we fix this quantity to a constant value. In order to do so, we express the KL divergence in terms of entropy.
\begin{equation}
\begin{split}
&D_{\mathit{KL}}[Q_{\phi}(\mathbf{z}|\mathbf{x}) \, \Vert \, P_{\theta}(\mathbf{z})] \\
&= -\mathbb{E}_{\mathbf{z} \sim Q_{\phi}(\mathbf{z} \vert \mathbf{x})}\left[\log \, P_{\theta}(\mathbf{z}) - \log \, Q_{\phi}(\mathbf{z} \vert \mathbf{x}) \right] \\
&= -\mathbb{E}_{\mathbf{z} \sim Q_{\phi}(\mathbf{z} \vert \mathbf{x})}\left[\log \, P_{\theta}(\mathbf{z}) \right] + \mathbb{E}_{\mathbf{z} \sim Q_{\phi}(\mathbf{z} \vert \mathbf{x})}\left[\log \, Q_{\phi}(\mathbf{z} \vert \mathbf{x}) \right] \\
&= H(Q_{\phi}(\mathbf{z} \vert \mathbf{x}), P_{\theta}(\mathbf{z})) - H(Q_{\phi}(\mathbf{z} \vert \mathbf{x}))
\end{split}
\label{kl_div_eq}
\end{equation}
where $H(Q_{\phi}(\mathbf{z} \vert \mathbf{X}), P_{\theta}(\mathbf{z}))$ represents the cross entropy between $Q_{\phi}(\mathbf{z}|\mathbf{X})$ and $P_{\theta}(\mathbf{z})$. It can be noted that we need to minimize the KL divergence term. Hence, if we assume that $H(Q_{\phi}(\mathbf{z} \vert \mathbf{x}))$ is constant, then we can drop this term during optimization (please refer to the next section for details of how $H(Q_{\phi}(\mathbf{z} \vert \mathbf{x}))$ is enforced to be constant). This lets us replace the KL divergence $D_{\mathit{KL}}[Q_{\phi}(\mathbf{z}|\mathbf{X}) \, \Vert \, P_{\theta}(\mathbf{z})]$ in the loss function with $H(Q_{\phi}(\mathbf{z} \vert \mathbf{X}), P_{\theta}(\mathbf{z}))$.
\begin{equation}
\begin{split}
&ELBO(\mathbf{x},\theta,\phi) \\
&= \mathbb{E}_{\mathbf{z}\sim Q_{\phi}(\mathbf{z}|\mathbf{x})}[\log P_\theta(\mathbf{x}|\mathbf{z})]
- H(Q_{\phi}(\mathbf{z} \vert \mathbf{x}), P_{\theta}(\mathbf{z})) \\
&= \mathbb{E}_{\mathbf{z}\sim Q_{\phi}(\mathbf{z}|\mathbf{x})}[\log P_\theta(\mathbf{x}|\mathbf{z})]
 + \mathbb{E}_{\mathbf{z} \sim Q_{\phi}(\mathbf{z} \vert \mathbf{x})}\left[\log \, P_{\theta}(\mathbf{z}) \right]
\end{split}
\label{kl_div_eqival}
\end{equation}
The choice of fixing the entropy of $Q_{\phi}(\mathbf{z} \vert \mathbf{x})$ is further justified via experiments in section \ref{expts}.

\subsection{Supervision using a Gaussian Mixture Prior}
\label{sup_gmm_prior}
In this section, we modify the above ELBO term for supervised learning by including the random variable $\mathbf{y}$ denoting labels. The following expression can be derived for the log-likelihood of the data.
\begin{equation}
\begin{split}
&\log (P_{\theta}(\mathbf{x},\mathbf{y})) = \mathbb{E}_{\mathbf{z}\sim Q_{\phi}(\mathbf{z}|\mathbf{x})}
[\log(P_{\theta}(\mathbf{x},\mathbf{y}|\mathbf{z}))] \\ &- D_{\mathit{KL}}[Q_{\phi}(\mathbf{z}|\mathbf{x}) || P_{\theta}(\mathbf{z})] + D_{\mathit{KL}}[Q_{\phi}(\mathbf{z}|\mathbf{x}) || P_{\theta}(\mathbf{z}|\mathbf{x},\mathbf{y})] 
\end{split}
\label{elbo_with_y_pre}
\end{equation}
Noting that $D_{\mathit{KL}}[Q_{\phi}(\mathbf{z}|\mathbf{x}) || P_{\theta}(\mathbf{z}|\mathbf{x},\mathbf{y})]\geq0$, and replacing $D_{\mathit{KL}}[Q_{\phi}(\mathbf{z}|\mathbf{x}) || P_{\theta}(\mathbf{z})]$ with $\mathbb{E}_{\mathbf{z} \sim Q_{\phi}(\mathbf{z} \vert \mathbf{x})}\left[\log \, P_{\theta}(\mathbf{z}) \right]$ by assuming $H(Q_\phi(\mathbf{z}|\mathbf{x}))$ to be constant (as shown in Eqn.~\ref{kl_div_eqival}), we get the following lower bound on the data likelihood.
\begin{equation}
\begin{split}
ELBO(\mathbf{x},\mathbf{y},\theta,\phi) = \mathbb{E}_{\mathbf{z}\sim Q_{\phi}(\mathbf{z}|\mathbf{x})} [\log(P_{\theta}(\mathbf{x},\mathbf{y}|\mathbf{z}))]  \\
+ \mathbb{E}_{\mathbf{z} \sim Q_{\phi}(\mathbf{z} \vert \mathbf{x})}\left[\log \, P_{\theta}(\mathbf{z}) \right]
\end{split}
\label{elbo_with_y}
\end{equation}
We choose our VAE to use a Gaussian mixture prior for the latent variable $\mathbf{z}$. We further choose the number of mixture components to be equal to the number of classes $k$ in the training data. The means of each of these components, $\boldsymbol{\mu}_{1},\boldsymbol{\mu}_{2},...,\boldsymbol{\mu}_{k}$ are assumed to be the one-hot encodings of the class labels in the latent space. It can be noted here that although this choice enforces the latent dimensionality to be $k$, it can be easily altered by choosing the means in a different manner. For example, means of all the mixture components can lie on a single axis in the latent space. Unlike usual VAEs, our encoder network outputs only the mean of $Q_\phi(\mathbf{z}|\mathbf{x})$. We use the reparameterization trick introduced by Kingma et al. \shortcite{kingma2013auto}, but sample the input $\boldsymbol{\epsilon}$ from $N(0,\Sigma_{constant})$ in order to enforce the entropy of $Q_\phi(\mathbf{z}|\mathbf{x})$ to be constant. Here, each mixture component corresponds to one class and $\mathbf{x}$ is assumed to be generated from the latent space according to $P_{\theta}(\mathbf{x}|\mathbf{z})$ irrespective of $\mathbf{y}$. Therefore, $\mathbf{x}$ and $\mathbf{y}$ become conditionally independent given $\mathbf{z}$, i.e. $\log(P_{\theta}(\mathbf{x},\mathbf{y}|\mathbf{z})) = \log(P_{\theta}(\mathbf{x}|\mathbf{z})) + \log(P_{\theta}(\mathbf{y}|\mathbf{z}))$. 
\begin{equation}
\begin{split}
&ELBO(\mathbf{x},\mathbf{y},\theta,\phi) \\ &= \mathbb{E}_{\mathbf{z}\sim Q_{\phi}(\mathbf{z}|\mathbf{x})} \left[\log(P_{\theta}(\mathbf{x}|\mathbf{z})) + \log(P_{\theta}(\mathbf{y}|\mathbf{z})) + \log \, P_{\theta}(\mathbf{z}) \right] \\
&= \mathbb{E}_{\mathbf{z}\sim Q_{\phi}(\mathbf{z}|\mathbf{x})} \left[\log(P_{\theta}(\mathbf{x}|\mathbf{z})) + \log(P_{\theta}(\mathbf{z}|\mathbf{y})) + \log \, P_{\theta}(\mathbf{y}) \right]
\end{split}
\label{elbo_with_y}
\end{equation}
Assuming the the classes to be equally likely, the final loss function for an input $\mathbf{x}^{(i)}$ with label $\mathbf{y}^{(i)}$ becomes the following.
\begin{equation}
\begin{split}
\mathcal{L}(\mathbf{x^{(i)}}, \mathbf{y}^{(i)}, \boldsymbol{\epsilon}) = ||\mathbf{x}^{(i)} &- g(f(\mathbf{x}^{(i)})+\epsilon) ||^2 \\
&+ \alpha|| f(\mathbf{x}^{(i)}) - \boldsymbol{\mu}_{\mathbf{y}^{(i)}} ||^2 %+ \beta \, \log\sum_{j=1}^{k} [exp (\boldsymbol{\mu_j} - f(\mathbf{x}^{(i)}))^2]
\end{split}
\label{final_loss}
\end{equation}
where the encoder is represented by $f$, the decoder is represented by $g$ and $\boldsymbol{\mu}_{\mathbf{y}^{(i)}}$ represents the mean of the mixture component corresponding to $\mathbf{y}^{(i)}$. $\alpha$ is a hyper-parameter that trades off between reconstruction fidelity, latent space prior and classification accuracy.
% It can be noted here that $\alpha$ and $\beta$ subsumes the variance of each of the mixture components of the prior. \pg{AL: attention here. Lines removed!! We no longer are changing $\theta$ of prior}

The label $\mathbf{y}$ for an input sample $\mathbf{x}$ can be obtained following the Bayes Decision rule.
\begin{equation}
\begin{split}
\arg\max_{\mathbf{y}}\,\, & P_{\theta}(\mathbf{y}|\mathbf{x}) = \arg\max_{\mathbf{y}} P_{\theta}(\mathbf{x,y})\\
&= \arg\max_{\mathbf{y}} \int_{\mathbf{z}} P_{\theta}(\mathbf{x,y}|\mathbf{z})P_{\theta}(\mathbf{z})d\mathbf{z} \\
&= \arg\max_{\mathbf{y}} \int_{\mathbf{z}} P_{\theta}(\mathbf{x}|\mathbf{z})P_{\theta}(\mathbf{y}|\mathbf{z})P_{\theta}(\mathbf{z})d\mathbf{z} \\
&= \arg\max_{\mathbf{y}} \int_{\mathbf{z}} P_{\theta}(\mathbf{z}|\mathbf{x})P_{\theta}(\mathbf{y}|\mathbf{z})P_{\theta}(\mathbf{x})d\mathbf{z} \\
&= \arg\max_{\mathbf{y}} \int_{\mathbf{z}} P_{\theta}(\mathbf{z}|\mathbf{x})P_{\theta}(\mathbf{y}|\mathbf{z})d\mathbf{z}
\end{split}
\label{classification_rule}
\end{equation}
% $\mathbf{y} = \arg\min_{c_i} || \boldsymbol{\mu}_{c_i} - f(\mathbf{x})||^2$. 
$P_{\theta}(\mathbf{z}|\mathbf{x})$ can be approximated by $Q_{\phi}(\mathbf{z}|\mathbf{x})$, i.e., the encoder distribution. This corresponds to the Bayes decision rule, in the scenario where there is no overlap among the classes in the input space, $\phi$ has enough variability and $Q_{\phi^*}(\mathbf{z}|\mathbf{x})$ is able to match $P_{\theta^*}(\mathbf{z}|\mathbf{x})$ exactly. 
% For further details and more general results, please refer to the appendix.

Semi-supervised learning follows automatically, by using the loss function in Eqn.~\ref{final_loss} for labeled samples, and the loss corresponding to Eqn.~\ref{kl_div_eqival} for unlabeled samples.

In order to compute the class label as defined in equation \ref{classification_rule}, we use a single sample estimate of the integration by simply using the mean of $Q_\phi(\mathbf{z}|\mathbf{x})$ as the $\mathbf{z}$ value in our experiments. This choice does not affect the accuracy as long as the mixture components representing the classes are well separated in the latent space.
% \AL{Intuitively, while VAEs try to match $Q_\phi(\mathbf{z}|\mathbf{x})$ with a constant variance prior, our network tries to maximize the ... can provide intuition claiming better/sharper reconstruction!}

% \subsection{Classification rule}

% \AL{The Bayes classification rule is given by $\arg\max_{\mathbf{y}} P(\mathbf{y}|\mathbf{x})$ Need to complete, and smoothly lead to appendix for the general case...} 

\subsection{Resisting adversarial attacks}
\label{res_adv_smpls}
In order to successfully reject adversarial samples irrespective of the method of its generation, we use thresholding at the encoder and decoder outputs. This allows us to reject any sample $\mathbf{x}$ whose encoding $\mathbf{z}$ has low probability under $P_\theta(\mathbf{z})$, i.e., if the distance between its encoding and the encoding of the predicted class label in the latent space exceeds a threshold value, $\tau_{enc}$ (since $P_\theta(\mathbf{z})$ is a mixture of Gaussians). We further reject those input samples which have low probability under $P_\theta(\mathbf{x}|\mathbf{z})$, i.e., if the reconstruction error exceeds a certain threshold, $\tau_{dec}$ (since $P_\theta(\mathbf{x}|\mathbf{z})$ is Gaussian). Essentially, a combination of these two thresholds ensures that $P_\theta(\mathbf{x}) = \int_\mathbf{z}P_\theta(\mathbf{x}|\mathbf{z})P_\theta(\mathbf{z})d\mathbf{z}$ is not low. 

Both $\tau_{enc}$ and $\tau_{dec}$ can be determined based on statistics obtained while training the model. In our experiments, we implement thresholding in the latent space as follows: we calculate the Mahalanobis distance between the encoding of the input and the encoding of the corresponding mixture component mean, and reject the sample if it exceeds the critical chi-square value ($3\sigma$ rule in the univariate case). Similarly, for $\tau_{dec}$, we use the corresponding value for the reconstructions errors.
% \footnote{\pg{Here we estimate mean reconstruction error by sample averaging and also find the standard deviation by assuming the reconstruction covariance take the form constant time identity matrix.}}. 
However, in general, any value can be assigned to these two thresholds, and they determine the risk to coverage trade-off for this selective classifier.

% In order to successfully fool this classifier, an adversary not only has to make sure that the latent representation of the adversarial sample lies within the $\tau_{enc}$ distance of the mean corresponding to the target class (or any incorrect class in a non-targeted case), but also that the reconstruction error lies within $\tau_{dec}$.

If the maximum allowed $L_{p}$ norm of the perturbation $\boldsymbol{\eta}$ is $\gamma$, then the adversary, trying to modify an input $\mathbf{x}$ from class $c_{1}$, must satisfy the following criteria.
\begin{enumerate}[leftmargin=1cm]
\setlength\itemsep{0em}
\item $\argmin_{c_i}|| f(\mathbf{x}+\boldsymbol{\eta}) - \boldsymbol{\mu}_{c_i} ||_{2} = c_{2}$ where $ c_{2}\neq c_{1}$
\item $ || f(\mathbf{x}+\boldsymbol{\eta}) - \boldsymbol{\mu}_{c_{2}} ||_{2} \leq \tau_{enc} $
%\item $\argmin_{i}||f(\mathbf{x}) - \boldsymbol{\mu}_{i} ||_{2} = c2$ where $ c1\neq %c2$
\item $||\boldsymbol{\eta}||_{p} \leq \gamma$
\item $|| (\mathbf{x}+\boldsymbol{\eta}) - g(f(\mathbf{x}+\boldsymbol{\eta}) + \boldsymbol{\epsilon}) ||_{2} \leq \tau_{dec}$ where $\boldsymbol{\epsilon} \sim N(0, \Sigma_{constant})$ 
\end{enumerate}
By the first three constraints, the encoding of $\mathbf{x}$ and $\mathbf{x}+\boldsymbol{\eta}$ must belong to different Gaussian mixture components in the latent space. However, constraint $4$ requires the distance between the reconstruction obtained from the encoding of $\mathbf{x}+\boldsymbol{\eta}$ to be close to $\mathbf{x}+\boldsymbol{\eta}$, i.e., close to $\mathbf{x}$ in the pixel space. This is extremely hard to satisfy because of the low probability of occurrence of holes in the latent space within $\tau_{enc}$ distance from the means.

Similarly, for the case of fooling samples, it can be argued that even if an attacker manages to generate a fooling sample which tricks the encoder, it will be very hard to simultaneously trick the decoder to reconstruct a similar image belonging to the rubbish class.

\subsection{Reclassification}
\label{reclassification}
Once a sample is detected as adversarial by either or both the thresholds discussed above, we attempt to find its true label using the decoder only. By definition of adversarial images, $\mathbf{x}_{adv} = \mathbf{x}_{org} + \boldsymbol{\eta}$, where $\mathbf{x}_{adv}$ is the adversarial image corresponding to the original image $\mathbf{x}_{org}$, and $||\boldsymbol{\eta}||_{p}$ is small. Hence, we can conclude that for any given image $\mathbf{x}$, $||\mathbf{x} - \mathbf{x}_{adv}||_p \approx ||\mathbf{x} - \mathbf{x}_{org}||_p$. Suppose $\mathbf{z}^*$ is given by Eqn.~\ref{finding_z}.
\begin{equation}
\mathbf{z}^* = \arg\min_{\mathbf{z}}{||g(\mathbf{z}) - \mathbf{x}_{org}||_p}
\label{finding_z}
\end{equation}
Following the argument stated above, we can approximate $\mathbf{z}^*\approx \mathbf{z}^*_{adv} = \argmin_\mathbf{z}{||g(\mathbf{z}) - \mathbf{x}_{adv}||_p}$. We can now find the label of the adversarial sample as $\arg\min_{c_i}{||\boldsymbol{\mu}_{c_i} - \mathbf{z}^*_{adv}||_2}$.
% \end{equation}
Essentially, for reclassification, we try to find the $\mathbf{z}$ in the latent space, which, when decoded, gives the minimum reconstruction error from the adversarial input. However, if Eqn.~\ref{finding_z} returns a $\mathbf{z}$ that lies beyond $\tau_{enc}$ from the corresponding mean, or if the reconstruction error exceeds $\tau_{dec}$, we conclude that the sample is a fooling sample and reject the sample. It can be noted here that if this network is deployed in a scenario where fooling samples are not expected to be encountered, one can choose not to reject samples during reclassification, thereby increasing coverage. Also, starting from a single value of $\mathbf{z}$ can cause the optimization process to get stuck at a local minimum. A better alternative is to run $k$ different optimization processes with $\mathbf{z} = \boldsymbol{\mu}_{1}, \boldsymbol{\mu}_{2}, \dots, \boldsymbol{\mu}_{k}$ as the initial values, and choose the $\mathbf{z}$ which gives minimum reconstruction error as $\mathbf{z}^*_{adv}$. Given enough compute power is available, these $k$ processes can be run in parallel. In our experiments, we follow these two strategies while reclassifying adversarial samples.

\section{Experiments}
\label{expts}
We verify the effectiveness of our network through numerical results and visual analysis on three different datasets - MNIST, SVHN and COIL-100. For different datasets, we make minimal changes to the hyper-parameters of our network, partly due to the difference in the image size and image type (grayscale/colored) in each dataset.

\begin{figure*}[t]
\begin{center}
\includegraphics[width=0.75\linewidth]{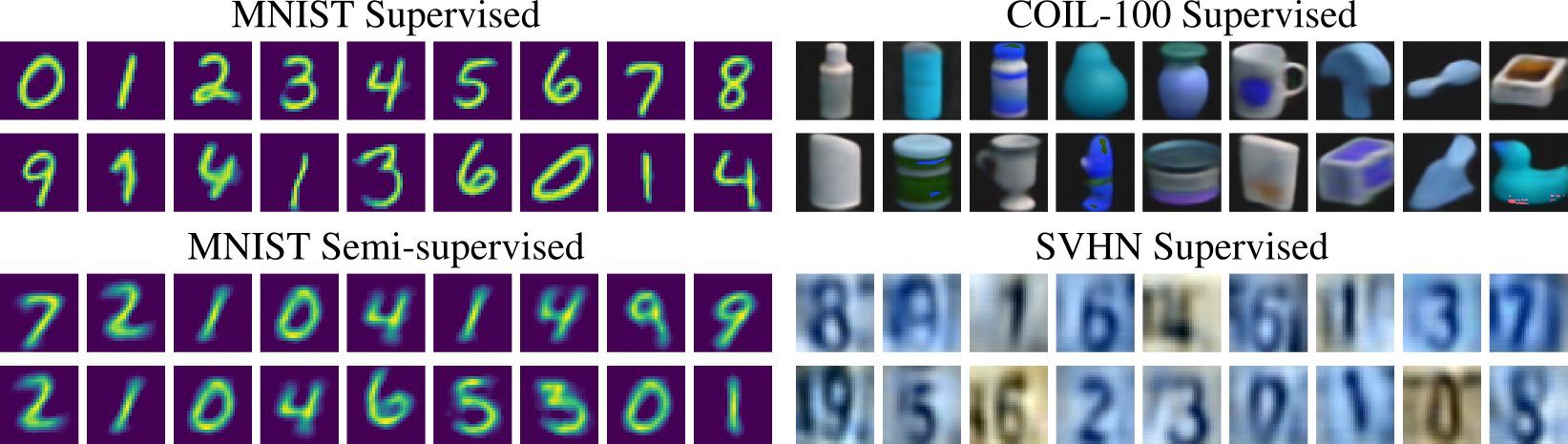}
\end{center}
\caption{Generated images from different classes of MNIST, COIL-100, SVHN.}
\label{generated_coil_data}
\end{figure*}

\subsubsection{Implementation details.}
We use an encoder network with convolution, max-pooling and dense layers to parameterize $Q_\phi(\mathbf{z}|\mathbf{x})$, and a decoder network with convolution, up-sampling and dense layers to parameterize $P_\theta(\mathbf{x}|\mathbf{z})$. We choose the dimensionality of the latent space to be the same as the number of classes for MNIST and COIL-100. However, noting that the size of images is larger for SVHN compared to MNIST, and also, because the dataset contains colored images, we choose the dimensionality of the latent space for SVHN as $32$ instead of $10$. The choice of means also varies slightly for this dataset, as we pad zeros to the one-hot encodings of the class labels to allow for the extra latent dimensions. The standard deviation of the encoder distribution is chosen such that the chance of overlap of the mixture components in the latent space is negligible and the classes are well separated. We use $1/3000$ as the variance for the MNIST dataset, and reduce this value as the latent dimensionality increases for the other datasets. We use the ReLU nonlinearity in our network, and sigmoid activation in the final layer so that the output lies in the allowed range $[0,1]$. We use the Adam\shortcite{kingma2014adam} optimizer for training. 
% For all our implementations, we rely on Keras \shortcite{chollet2015keras}. 
% Since achieving state-of-the-art-accuracy is not our primary goal, we do not perform much hyper-parameter tuning for improving the accuracy. 

\begin{center}
\begin{table*}[t!]
  \centering
  \begin{tabular}{cccc c c c c c c c}
      \hline
     &  \multicolumn{5}{c}{\underline{\qquad \qquad \quad \qquad Supervised\qquad \qquad \qquad}\qquad \quad} & \multicolumn{4}{c}{\underline{\qquad \qquad \quad Semi-supervised \quad\qquad \qquad}\qquad \qquad \qquad}\rule{0pt}{2.6ex}\\
      \vspace{-.15cm}
      & \multicolumn{2}{c}{Without} & \multicolumn{3}{c}{With} & \multicolumn{1}{c}{Without} & \multicolumn{3}{c}{With} \rule{0pt}{2.6ex}\\
      & \multicolumn{2}{c}{\underline{\quad thresholding\quad}} & \multicolumn{3}{c}{\underline{\qquad\qquad thresholding \qquad\quad}} & \multicolumn{1}{c}{\underline{thresholding}} & \multicolumn{3}{c}{\underline{\qquad\quad\quad thresholding\quad\qquad\qquad}} \rule{0pt}{2.6ex}\\
\vspace{.1cm}
      Dataset & SOTA & Accuracy & Accuracy & Error & Rejection & Accuracy & Accuracy & Error & Rejection\rule{0pt}{2.6ex}\\
%       \hline
      MNIST & 99.79\% & 99.67\% & 97.97\% & 0.22\%  & 1.81\% & 99.1\% & 98.17\% & 0.52\% & 1.31\%\\
      % \hline
      SVHN & 98.31\% & 95.06\% & 92.80\% & 4.58\% & 2.62\%  & 86.42\% & 83.54\% & 13.64\% & 2.82\%\\
      % \hline
      COIL-100 & 99.11\% & 99.89\% & 98.40\% & 0\% & 1.60\% & - & - & - & -\\
      \hline
  \end{tabular}
%   \vspace{0.25cm}
  \caption{Comparison between the performance of the state-of-the-art (SOTA) models and our model. We show that our method, even without much fine tuning focused on achieving classification accuracy, is competitive with the SOTA. MNIST SOTA is as reported by \shortcite{wan2013regularization}, SVHN SOTA is as given by \shortcite{lee2016generalizing} and the SOTA for COIL-100 is given by \shortcite{wu2015kernel}.}
  \label{Comp_normal_classifier}
\end{table*}
\end{center}

\subsubsection{Qualitative evaluation.}
Since our algorithm relies upon the reconstruction error between the generated and the original samples, we first show a few randomly chosen images generated by the network (for both supervised ad semi-supervised scenarios) corresponding to test samples of different classes from the three datasets in Figure \ref{generated_coil_data}.

\subsubsection{Numerical results.}
In Table \ref{Comp_normal_classifier}, we present the accuracy, error and rejection percentages obtained by our method with and without thresholding. For semi-supervised learning, we have taken $100$ randomly chosen labeled samples from each class for both MNIST and SVHN during training. 
% At this point, it should be noted that our main goal is not to achieve state of the art classification accuracy, but to build classifiers that are robust against adversarial attacks. Hence, we do not perform much hyper-parameter tuning for improving the accuracy achieved by our network. 
It is important to note here that the SOTA for COIL-100 was obtained on a random train-test split of the dataset, and hence, the accuracy values are not directly comparable.

\subsubsection{Adversarial attacks on encoder.}
\begin{figure}[h]
\centering
\includegraphics[width=0.45\textwidth]{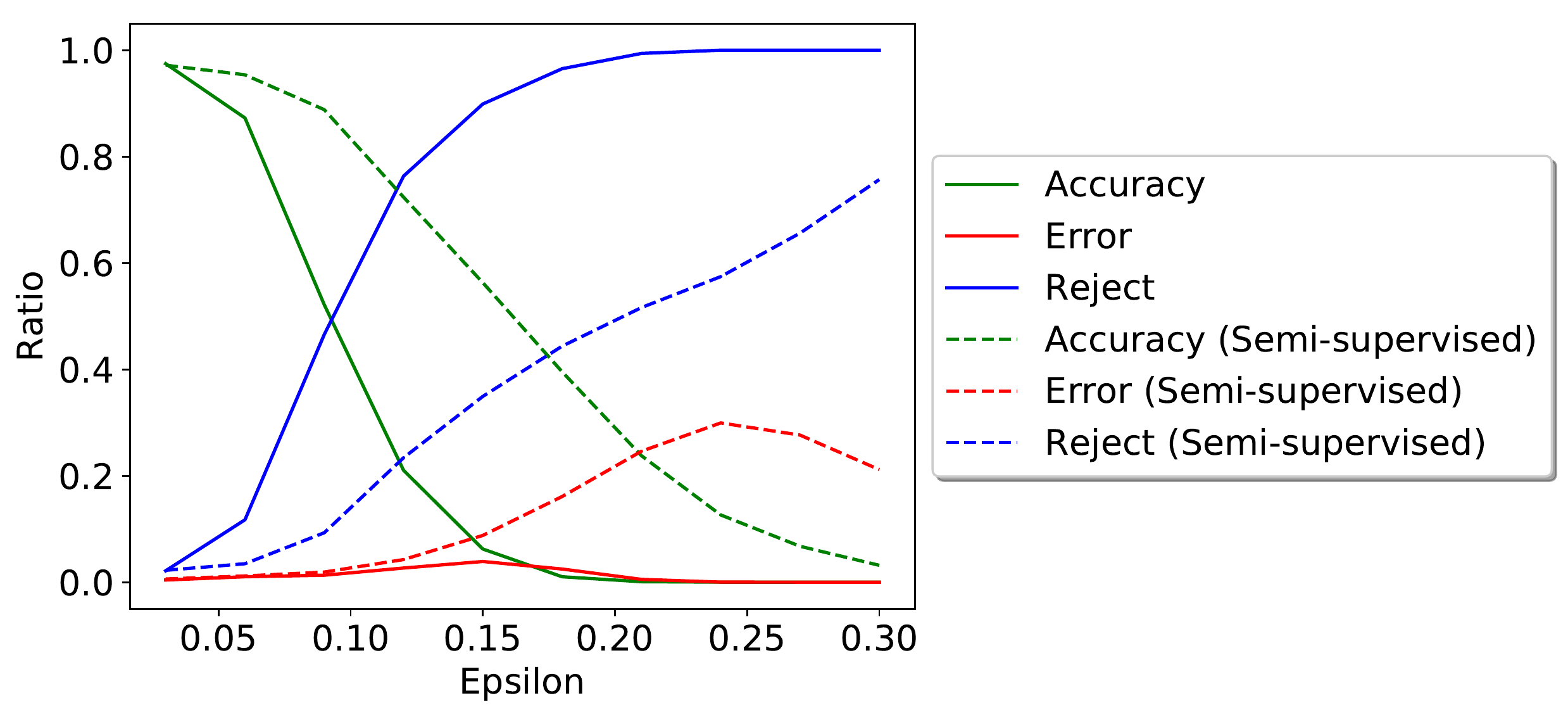}
%     \subfigure[Adversarial samples for MNIST dataset. Each odd row represent the initial class and each even row represent the reconstructed class of its previous row. The visual similarity indicates correct classification and and the visual mismatch indicates rejection.]{
%       \raisebox{-1\height}{\includegraphics[width=0.48\textwidth]{drawing.pdf}}
%       \label{full_mnist_adv_smpls}}\qquad
%     \subfigure[All methods fail to fool the generative model. This results from two facts. 1. the generator maps a low dimensional input and hence is much less likely to have regions that produce data points that are outside the data manifold. 2. The attack methods do not explicitly target the generator. This establishes non-transferability of adversarial samples to our model from arbitrary CNNs.]{
%         \raisebox{-1\height}{\includegraphics[height=200pt, width=0.48\linewidth]{attack_vs_def.pdf}}
%         \label{all_ineffective}}
  \caption{We run FGSM with varying $\epsilon$ on the models trained on MNIST data in both supervised and semi-supervised scenarios. Although the error rate is higher for the semi-supervised network, the rejection ratio rises monotonically for both networks with increasing $\epsilon$, and the error rate for the supervised model stays below 5\%.}
  \label{fgsm_fig}
\end{figure}

We use the encoder part of the network trained on the MNIST dataset to generate adversarial samples using the \textit{Fast Gradient Sign Method (FGSM)} with varying $\epsilon$ values \shortcite{goodfellow2014explaining}. The corresponding results are shown in Figure \ref{fgsm_fig}. The behavior is as desired, i.e., with increasing $\epsilon$, percentage of misclassified samples rises to a maximum value of only $3.89\%$ and then decreases, while the accuracy decreases monotonically and the rejection percentage increases monotonically. Similar results are obtained for the semi-supervised model, as shown in Figure \ref{fgsm_fig}, although the maximum error rate is higher in this case. We further tried the FGSM attack from the Cleverhans library \shortcite{papernot2017cleverhans} with the default parameters on the SVHN and COIL-100 datasets, and all the generated samples were rejected by the models after thresholding. Similarly, we generated adversarial samples for all three datasets using stronger attacks from Cleverhans with default parameter settings, including the Momentum Iterative Method \shortcite{dongboosting} and Projected Gradient Descent \shortcite{madry2017towards}. In these cases as well, all generated adversarial samples were successfully rejected by thresholding.

This indicates that since all these attacks lack knowledge of the decoder network, they only manage to produce samples which fool the encoder network, but are easily detected at the decoder output. From this set of experiments, we conclude that the only effective method of attacking our model would be to design a complete white-box attack that has knowledge of the decoder loss as well, as well as the two thresholds. Further, since we do not use any form of gradient obfuscation in our defense mechanism, a complete white-box attacker would represent a strong adversary.
\subsubsection{White-box adversarial attack.}
\label{whitebox_attc_sec}
\begin{figure*}
  \centering
  \setlength\tabcolsep{2pt}
  \begin{tabular}{ccc}
  	 Adversarial Samples (MNIST) & Adversarial Samples (COIL) & Fooling Samples\\
     \includegraphics[width=0.325\linewidth]{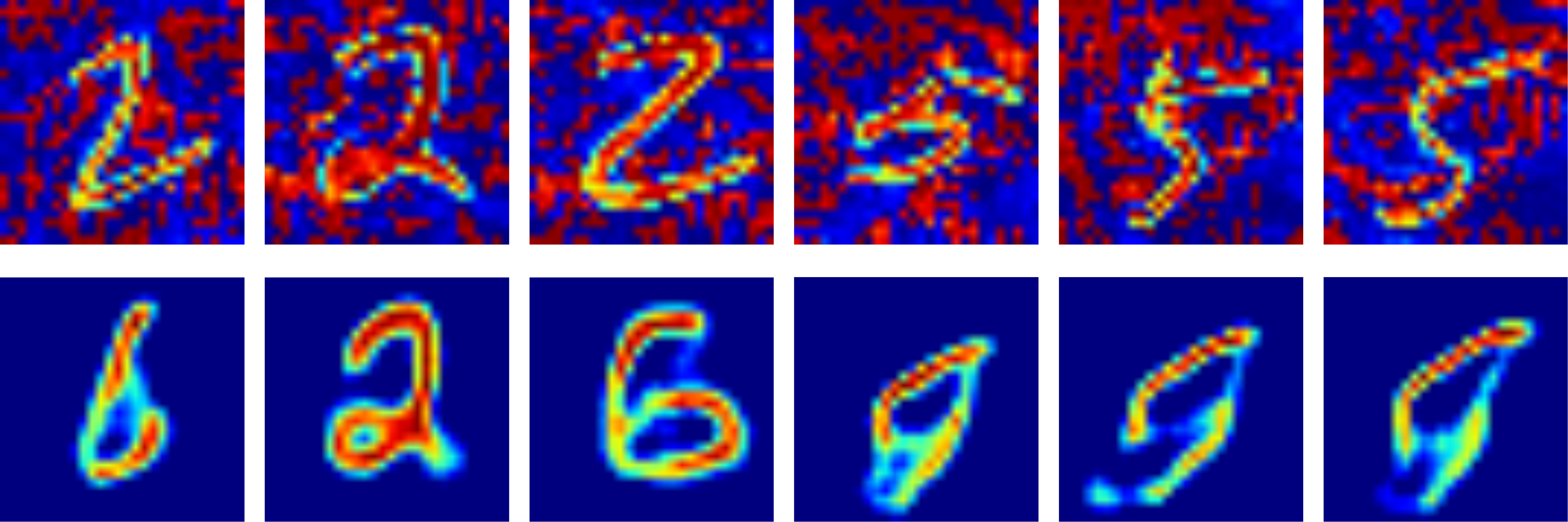} &
     \includegraphics[width=0.325\linewidth]{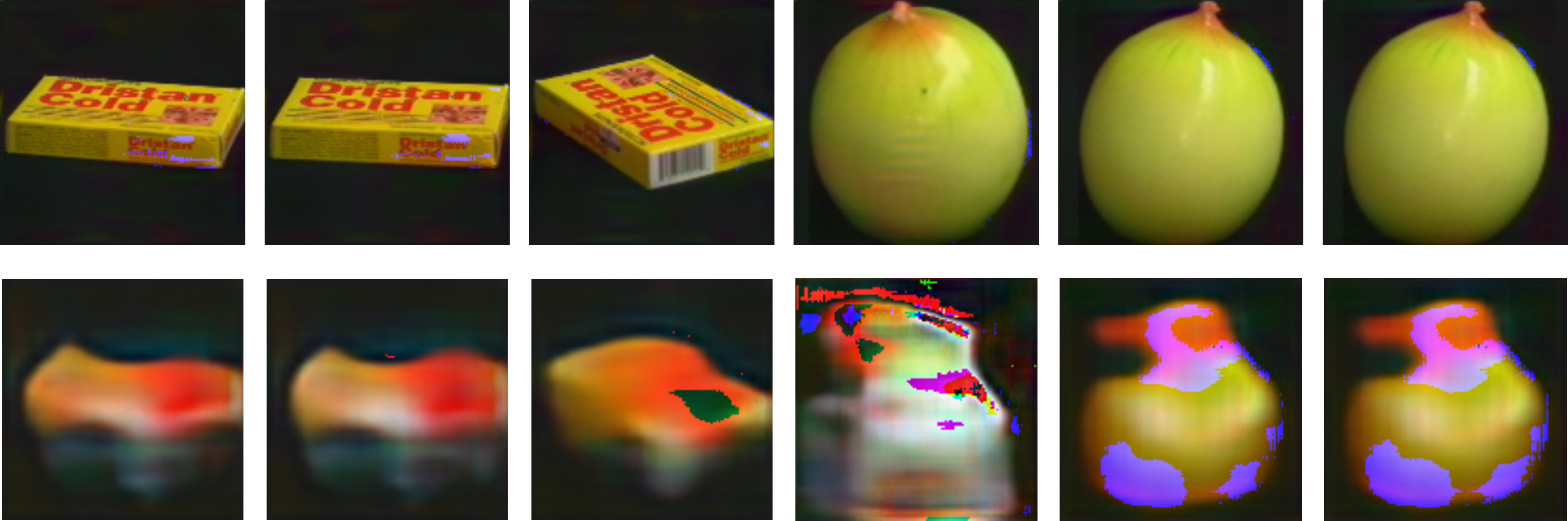} &
     \includegraphics[width=0.325\linewidth]{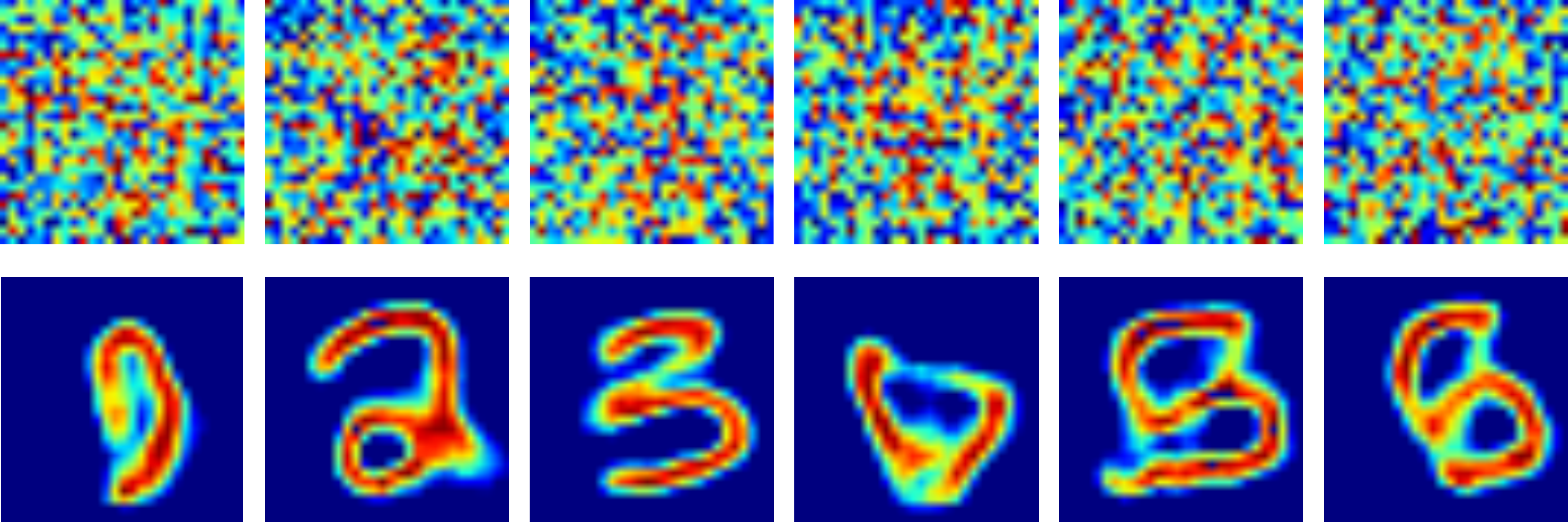}\\
     (a) & (b) & (c)
    \end{tabular}
  \caption{White box attack on MNIST and COIL dataset. (a) Targeted attack on MNIST. (b) Targeted attack on COIL. (c) Targeted fooling sample attack on MNIST. The first row represents the images to which the white-box attack converged, and the second row represents the corresponding reconstructed images.}
  \label{fooling_and_adv_samples}
\end{figure*}

We present the results for completely white-box targeted attack on our model for the COIL-100 and MNIST datasets in figures \ref{fooling_and_adv_samples}a and \ref{fooling_and_adv_samples}b. Here, the adversary has complete knowledge of the encoder, the decoder, as well as the rejection thresholds. The results shown correspond to random samples from the first two classes of objects for the COIL-100 dataset, and the classes $2$ and $5$ for MNIST dataset. We perform gradient descent on the adversarial objective as given in Eqn.~\ref{ad_obj}. The target class is set to $6$ for MNIST images from class $2$, $9$ for MNIST images from class $5$, and the class other than that of the source image for the COIL-100 images.
\begin{equation}
\begin{split}
&\argmin_{\boldsymbol{\eta}}{\mathcal{L}_{adv}} = \\
& \qquad \qquad \argmin_\mathbf{\eta} [((||\mathbf{x_o} + \boldsymbol{\eta} - g(f(\mathbf{x_o} + \boldsymbol{\eta}))||^2)/\tau_{dec})^a \\ 
& + ((\mathbf{\mu_t} - f(\mathbf{x_o} + \mathbf{\eta}))\Sigma_t(\mathbf{\mu_t} - f(\mathbf{x_o} + \boldsymbol{\eta}))/\tau_{enc})^b + ||\boldsymbol{\eta}||^2]
\label{ad_obj}
\end{split}
\end{equation}
where $\mathbf{x_o}$ is the original image we wish to corrupt, $\boldsymbol{\mu_t}$ is the mean of target class, $\boldsymbol{\eta}$ is the noise added, $f, g$ are the encoder and decoder respectively, and $\Sigma_t$ denotes target class covariance in latent space. $a>1$ and $b>1$ represent constant exponents which ensure that the adversarial loss grows steeply when the two threshold values are exceeded. Essentially, we aim for low reconstruction error and small change in the adversarial image while moving its embedding close to the target class mean. $\boldsymbol{\eta}$ is initialized with zeros. 

We also ran the white box attack on $100$ randomly sampled images from each of the $10$ classes for MNIST and SVHN, by setting each of the $9$ other classes as the target class. The samples generated by optimizing the adversarial objective in each of these cases were either correctly classified or rejected.

\subsubsection{Fooling images.}
We take $100$ images sampled from the uniform distribution as inputs and optimize the white-box fooling attack objective given by Eqn. \ref{fooling_obj}, with each of the classes from the MNIST and SVHN datasets as the target classes. In Figure \ref{fooling_and_adv_samples}c, we visualize some of the images to which the attack converged and their reconstructions for the MNIST dataset, with the target classes $1, 2, \dots, 6$.
\begin{equation}
\begin{split}
\argmin_{\boldsymbol{\eta}}&{\mathcal{L}_{fool}} = \argmin_{\boldsymbol{\eta}}[(||\boldsymbol{\eta} - g(f(\boldsymbol{\eta}))||^2/\tau_{dec})^a \\
&+ ((\boldsymbol{\mu_t} - f(\boldsymbol{\eta}))\Sigma_t(\boldsymbol{\mu_t} - f(\boldsymbol{\eta}))/\tau_{enc})^b]
\label{fooling_obj}
\end{split}
\end{equation}
Here, $\boldsymbol{\eta}, a, b, f$, $g$, $\Sigma_t$ and $\boldsymbol{\mu_t}$ are as described in sec. \ref{whitebox_attc_sec}. 

It has been shown that fooling samples are extremely easy to generate for state-of-the-art classifier networks \shortcite{goodfellow2014explaining,nguyen2015deep}. Our technique, by design, gains resilience against such attacks as well. Since by definition, a fooling sample cannot look like a legitimate sample, it can not have small pixel space distance with any real image. This is exactly what can be noticed in the results in Figure \ref{fooling_and_adv_samples}c, where reconstruction errors are very high. Hence, most of the images to which this attack converges are rejected at the decoder, although they had managed to fool the encoder when considered in isolation. For the few cases where the images are not rejected, we observe that the attack method actually converged to a legitimate image of the target class.
% We further argue that since this white-box adversarial attack can not find a successful solution by optimizing the given objective, it is extremely unlikely that a directed random sampling will accomplish the same, which is one of the attack methods suggested by Goodfellow et al. \shortcite{goodfellow2014explaining}.

% \begin{figure}
%   \centering
%   \setlength\tabcolsep{-2pt}
%   \begin{tabular}{cc}
%       \begin{tabular}{c}
%           \begin{tabular}{c}
%              \includegraphics[width=0.48\linewidth]{fooling_smpls.pdf}\\
%              Reconstruction Error
%           \end{tabular}\\
%           \setlength\tabcolsep{6pt}
%           \begin{tabular}{c|c|c|c|c|c}
%         %     \hline
%             0.176 & 0.192 & 0.190 & 0.190 & 0.178 & 0.179 \\
%         %     \hline
%           \end{tabular}
%       \end{tabular} &
%       \begin{tabular}{c}
%           \setlength\tabcolsep{6pt}
%           \begin{tabular}{c}
%              \includegraphics[width=0.48\linewidth]{fooling_smpls.pdf}\\
%              Reconstruction Error
%           \end{tabular}\\
%           \setlength\tabcolsep{6pt}
%           \begin{tabular}{c|c|c|c|c|c}
%         %     \hline
%             0.176 & 0.192 & 0.190 & 0.190 & 0.178 & 0.179 \\
%         %     \hline
%           \end{tabular}
%       \end{tabular}
%   \end{tabular}
%   \caption{Generating fooling samples for MNIST classifier. It can be noticed that the reconstructed samples are natural and clearly, they are very far apart in the pixel space as compared to the $3\sigma$ bound of reconstruction error, which is $0.0331$}
%   \label{foolingsamples}
% \end{figure}

\subsubsection{Reclassifying Adversarial samples.}
In this section we present the performance of our reclassification technique. Although one could have used our decoder network to perform both ``ordinary'' and ``adversarial'' sample classification using Eqn.~\ref{finding_z}, but this process involves an iterative optimization. Hence, we only use it for the detected adversarial samples. The results are summarized in Table \ref{reclassification_acc}.
\begin{table}[H]
\centering
\begin{tabular}{c c c c c c c c c c}
\hline
$\epsilon$ & 0.06 & 0.12 & 0.18 & 0.24 & 0.30 \\
Accuracy & 97\% & 93\% & 91\% & 87\% & 87\% \\
\hline
\end{tabular}
\caption{We present the reclassification accuracy for samples generated using FGSM on the MNIST dataset.}
\label{reclassification_acc}
\end{table}
Following the same reclassification scheme, we also find that the method is able to correctly classify rejected test samples, thereby improving the overall accuracy achieved by the proposed method. For example, among the 181 samples rejected by the supervised model for the MNIST test dataset (as per Table \ref{Comp_normal_classifier}), 110 samples are now correctly classified, improving the accuracy to 99.07\%.
\subsubsection{Entropy of $Q_{\phi}(\mathbf{z} \vert \mathbf{x})$.}
\label{const_entropy}
\begin{figure}[h]
\centering
\includegraphics[width=0.3\textwidth]{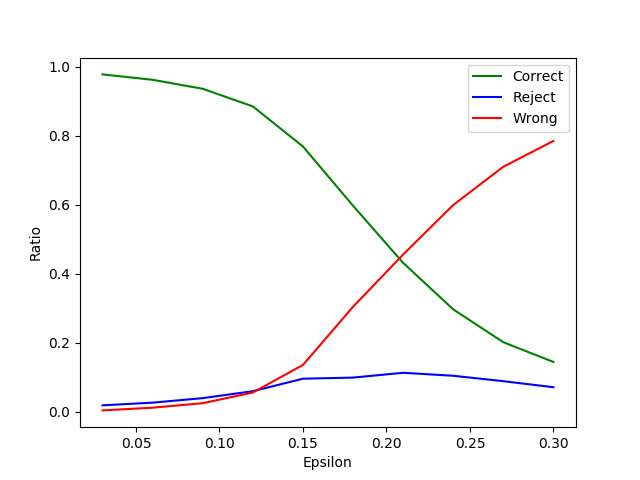}
\caption{We run FGSM with varying $\epsilon$ on the model with variable encoder distribution entropy, trained on MNIST data. The rejection rate stays low in this case, while the error rate increases with increasing $\epsilon$.}
  \label{fgsm_variable}
\end{figure}

To compare the performance of the proposed network with the corresponding network with variable entropy of $Q_{\phi}(\mathbf{z} \vert \mathbf{x})$, we ran experiments by letting $H(Q_{\phi}(\mathbf{z} \vert \mathbf{x}))$ to be variable, and keeping all other parameters same. We tried the FGSM attack against the encoder of the model thus obtained, and observed that the adversarial sample detection capability of the network reduces drastically. This is justified by the fact that the reconstructions tend to be blurry in this case, thereby leading to a high reconstruction threshold. The results are shown in figure \ref{fgsm_variable}.
In order to further study the difference between the two cases, we train both variants of the network on the CelebA dataset, and observe that the “Fr\'echet Inception Distance (FID) \shortcite{heusel2017gans} score is significantly better for the model with a constant $H(Q_{\phi}(\mathbf{z} \vert \mathbf{x}))$ (50.4) than the one with variable $H(Q_{\phi}(\mathbf{z} \vert \mathbf{x}))$ (58.3). The FID scores are obtained by randomly sampling $10,000$ points from the latent distribution, and comparing the distribution of the images generated from the these points with the training image distribution. 

\section{Discussion}
In this work, we have successfully demonstrated how a generative model can be used to gain defensive strength against adversarial attacks on images of relatively high resolution (128x128 for the COIL-100 dataset for example). However, the proposed network is limited by the generative capability of VAE based architectures, and thus, might not scale effectively to ImageNet scale datasets \shortcite{imagenet_cvpr09}. In spite of this fact, keeping the underlying principles for adversarial sample detection and reclassification as described in this work, recent advances in invertible generative models such as Glow \shortcite{kingma2018glow} can be exploited to scale to more complex datasets. Further, as discussed earlier, the problem of defending against adversarial attacks still remains an unsolved problem even for datasets with more structured images. Hence our method can be used for practical applications such as secure medical image classification \shortcite{finlayson2018adversarial}, biometrics identification, etc.

Human perception involves both discriminative and generative capabilities. Similarly, our work proposes a modification to VAEs to incorporate discriminative ability, besides using its generative ability to gain robustness against adversarial samples. The input space dimensionality (to the decoder) is drastically smaller compared to the input space dimensionality of image classifiers. Hence, it is much easier to attain dense coverage in the latent space, thereby minimizing the possibility of the occurrence of holes, leading to defensive capability against both adversarial and fooling images. With our construct, selective classification and semi-supervised learning become feasible under the same framework. A possible direction of future research would be to study how effectively the proposed approach can be scaled to more complex datasets by using recently proposed invertible generative modeling techniques.

\section{Acknowledgement}
We are extremely grateful to Mr. Arnav Acharyya for his invaluable contribution to the discussions that helped shape this work.
\bibliographystyle{aaai}
\bibliography{ResistingAdvAtack}
\end{document}